\documentclass[10pt,twocolumn,letterpaper]{article}

\usepackage{cvpr_draft}
\usepackage[T1]{fontenc}
\usepackage[utf8]{inputenc}
\usepackage{amsmath,amssymb,mathtools}
\usepackage{graphicx,booktabs,array,multirow,tabularx}
\usepackage{microtype}
\usepackage{enumitem}
\usepackage{url}
\usepackage{listings}
\usepackage{algorithm}
\usepackage[noend]{algpseudocode}
\usepackage{placeins}
\usepackage[colorlinks=true,citecolor=blue!45!black,linkcolor=blue!45!black,urlcolor=blue!45!black]{hyperref}
\usepackage[capitalize,nameinlink]{cleveref}
\definecolor{pendingblue}{HTML}{285C87}
\definecolor{paperteal}{HTML}{087F8C}
\newcommand{\method}{\mbox{\normalfont\scshape KnowBody}}

\newcommand{\tightpara}[1]{\noindent\textbf{#1}\quad}
\newcommand{\unknown}{\mathsf{unknown}}

\setlist[itemize]{leftmargin=*,topsep=3pt,itemsep=2pt,parsep=0pt}
\setlist[enumerate]{leftmargin=*,topsep=3pt,itemsep=2pt,parsep=0pt}
\lstdefinestyle{appendixblock}{
  basicstyle=\ttfamily\fontsize{7.3}{8.1}\selectfont,
  breaklines=true,
  breakatwhitespace=false,
  columns=fullflexible,
  frame=single,
  rulecolor=\color{black!18},
  backgroundcolor=\color{black!2},
  showstringspaces=false,
  keepspaces=true,
  aboveskip=5pt,
  belowskip=8pt,
  xleftmargin=2pt,
  xrightmargin=2pt
}
\newcolumntype{L}[1]{>{\raggedright\arraybackslash}p{#1}}
\newcolumntype{Y}{>{\raggedright\arraybackslash}X}
\makeatletter
\providecommand{\theHALG@line}{\arabic{ALG@line}}
\newcommand{\setupappendixnumbering}{%
  \@addtoreset{figure}{section}%
  \@addtoreset{table}{section}%
  \@addtoreset{algorithm}{section}%
  \@addtoreset{lstlisting}{section}%
  \renewcommand{\thefigure}{\Alph{section}\arabic{figure}}%
  \renewcommand{\thetable}{\Alph{section}\arabic{table}}%
  \renewcommand{\thealgorithm}{\Alph{section}\arabic{algorithm}}%
  \renewcommand{\thelstlisting}{\Alph{section}\arabic{lstlisting}}%
  \renewcommand{\theHfigure}{appendix.\Alph{section}.\arabic{figure}}%
  \renewcommand{\theHtable}{appendix.\Alph{section}.\arabic{table}}%
  \renewcommand{\theHalgorithm}{appendix.\Alph{section}.\arabic{algorithm}}%
  \renewcommand{\theHlstlisting}{appendix.\Alph{section}.\arabic{lstlisting}}%
  \renewcommand{\theHALG@line}{\thealgorithm.\arabic{ALG@line}}%
}
\makeatother
\crefname{algorithm}{algorithm}{algorithms}
\Crefname{algorithm}{Algorithm}{Algorithms}
\crefname{lstlisting}{listing}{listings}
\Crefname{lstlisting}{Listing}{Listings}
\graphicspath{{figures/}}
\hypersetup{pdftitle={Know Your Body: A Continual Embodiment Harness for Direct VLM Control},pdfsubject={Body-grounded VLM control; four-task comparison and continual trial sequences}}

\title{%
Know Your Body: A Harness\\[-0.05em]
for Direct and Self-Improving Robot Control with VLMs\\[0.35em]
{\small\normalfont\sffamily
\href{https://loule0-0.github.io/KnowBody/}
{\textcolor{blue!65!black}{https://loule0-0.github.io/KnowBody/}}}%
}

\author{
Zeyu Lou\textsuperscript{1,2} \quad
Yanhong Zeng\textsuperscript{2} \quad
Yong~Wang\textsuperscript{2,3} \quad
Chenyang Si\textsuperscript{1,\textdagger}\\[3pt]
{\small
\textsuperscript{1}Nanjing University \quad
\textsuperscript{2}Ant Group \quad
\textsuperscript{3}Zhejiang University}
}

\date{}

\begin{document}

\twocolumn[\maketitle]

\begingroup
\makeatletter
\renewcommand{\@makefntext}[1]{%
  \noindent
  \makebox[1em][l]{\@makefnmark}#1%
}
\makeatother

\renewcommand{\thefootnote}{\textdagger}
\footnotetext{%
\parbox[t]{\dimexpr\linewidth-1em\relax}{%
\raggedright
Corresponding author.\\[-1pt]
Emails: \nolinkurl{zeyu.lou.mail@gmail.com},
\nolinkurl{chenyang.si@nju.edu.cn}.
}}
\endgroup

\raggedbottom

\suppressfloats[t]

\begin{abstract}
A general-purpose vision--language model can understand a task goal without knowing how a particular robot's motion and functional parts produce the intended effect. We introduce \textbf{KnowBody}, a harness that makes these action-relevant body relations explicit, queryable, and revisable while keeping the model weights frozen. Initialized from one off-task trajectory, a partial body model guides action selection and the interpretation of past interactions. New evidence refines the model, and knowledge dependent on revised body estimates is rechecked before reuse. Across 32 fixed-budget trials on four real-robot tasks, initialized KnowBody achieves 75\% completion versus 25\% for the native harness and requires fewer planner rounds on successful trials in tasks completed by both. With persistent updates enabled, planner rounds decrease by 29--53\% from the first to the fifth recorded success.
\end{abstract}

\begin{figure}[t]
\centering
\includegraphics[width=\columnwidth]{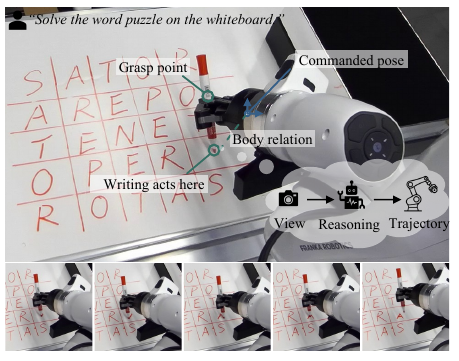}
\caption{\textbf{Knowing the task is not knowing the body.} A VLM is able to recognize what to write, but the robot receives an end-effector pose while writing acts at the held tip. \method\ makes this command--effect relation explicit for action decision and outcomes interpretation.}
\label{fig:teaser}
\end{figure}
\begin{figure*}[t]
\centering
\includegraphics[width=\textwidth]{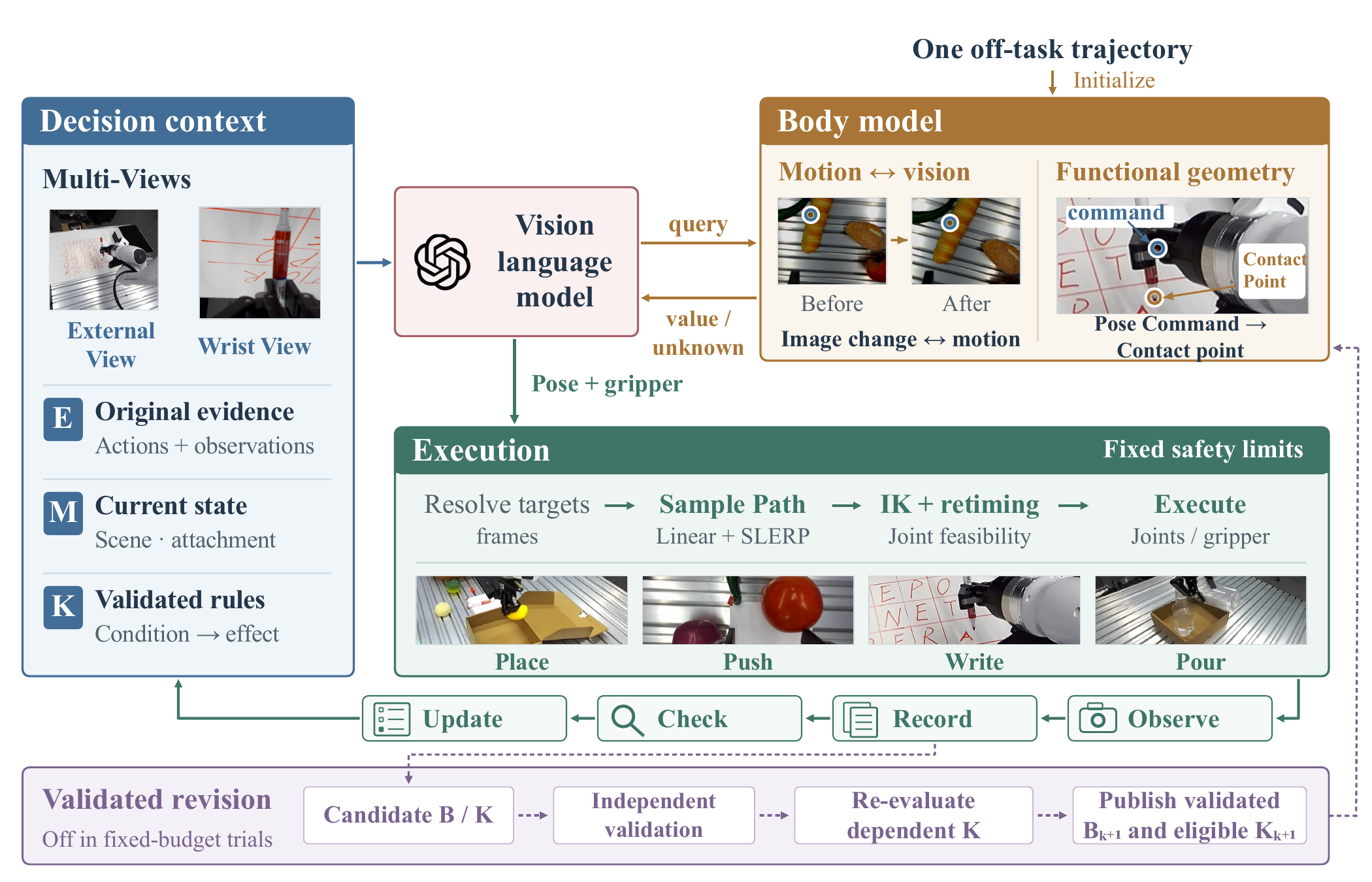}
\caption{\textbf{KnowBody separates online state feedback from validated cross-episode revision.} At each decision, the VLM receives the task, current observations, current state $M$, eligible rules $K$, and selected original evidence $E$. It queries the partial body model, issues waypoints and gripper commands, and receives execution feedback through Observe $\rightarrow$ Record $\rightarrow$ Check $\rightarrow$ Update $M$; only $M$ returns directly to the next decision. Qualified evidence enters the revision lane, where candidate $B/K$ changes are independently validated and rules dependent on a changed $B$ are re-evaluated. Dashed purple arrows publish validated $B_{k+1}$ to the body model and eligible $K_{k+1}$ to the decision context for later episodes. One off-task trajectory initializes supported relations in $B_0$.}
\label{fig:harness}
\end{figure*}
\section{Introduction}
\label{sec:introduction}

General-purpose vision--language models (VLMs) offer an appealing route to robot manipulation: their broad visual and task knowledge can support decisions across tasks, while a harness connects those decisions to robot observations, action tools, and execution feedback. This approach makes it possible to attempt new tasks without first training a task-specific policy, and offers a way to bring advances in general-purpose models into robot control. Yet broad task knowledge does not establish how a particular robot's motions and functional parts produce the intended physical effects.

Consider writing with a grasped pen: the task requires contact at the pen tip, whereas the robot receives end-effector pose commands. Choosing those commands requires knowing how the held tip relates to the commanded pose. Observations and interaction histories provide evidence about such relations; using that evidence for control requires identifying what is supported, applying it to the next action, and revising it when new observations reveal a mismatch. The central challenge is therefore to connect action selection and learning through explicit body knowledge that a general-purpose VLM can use and that subsequent interactions can test and refine.

We introduce \textbf{KnowBody}, a harness that connects a  VLM's action decisions to a partial, queryable body model. Before acting, the VLM uses relevant motion and functional-geometry relations, together with current observations and knowledge of previous action outcomes, to choose concrete robot commands. Execution feedback and subsequent observations then serve two purposes: assessing task progress and testing the body relations used in the decision. Evidence from these interactions supports the refinement of body estimates and the accumulation of conditional knowledge about action outcomes, which inform later decisions. The body model thus provides a shared reference for choosing actions and learning from their consequences, while the VLM remains responsible for task reasoning and action selection.

A single off-task trajectory initializes the body model with relations supported by its recorded actions and observations. Subsequent interactions provide evidence for refining body estimates and acquiring conditional knowledge about action outcomes. Validated updates are retained across episodes, allowing later decisions to draw on knowledge accumulated during earlier tasks. When a body estimate changes, interaction knowledge that depends on it is rechecked before reuse.

We evaluate KnowBody on four real-robot tasks: duck placement, apple pushing, writing, and pouring. Across 32 fixed-budget trials with the same frozen VLM and cross-episode updates disabled, KnowBody achieves 75\% completion versus 25\% for the native Codex harness. On the three tasks completed by both methods, successful KnowBody trials require 17--44\% fewer planner rounds on average. With persistent updates enabled, rounds decrease by 29--53\% from the first to the fifth recorded success across the four tasks.

\section{Related Work}
\label{sec:related}

\tightpara{General models as robot decision makers.}
SayCan grounds language-model proposals with skill-value estimates~\citep{ahn2023saycan}, while PaLM-E incorporates continuous sensor observations into a trained embodied language model~\citep{driess2023palme}. Code as Policies generates programs over perception and control APIs~\citep{liang2023code}; CaP-X examines how primitive abstraction, visual grounding, and execution feedback affect coding-agent manipulation~\citep{fu2026capx}. PIVOT iteratively selects image-space actions~\citep{nasiriany2024pivot}, and VLMPC uses VLM-proposed sequences in a predictive-control loop~\citep{zhao2024vlmpc}.

Show-Harness interprets discrete semantic motion units as local robot actions, with multi-view guidance, proprioception, and action chunking~\citep{chen2026showharness}. Agent as Policy (AGP) uses calibrated observations, geometric queries, and runtime programs to generate and revise motion~\citep{jia2026agentpolicy}. GPT-Policy compiles demonstrations, goal images, interaction history, and execution feedback into context for a fixed VLM acting through a constrained controller~\citep{cheng2026gptpolicy}. \method\ builds on these direct-control interfaces, focusing on body evidence from one off-task trajectory and on experience whose validity depends on that evidence.

\tightpara{Trained visuomotor policies and VLAs.}
Diffusion Policy learns multimodal action sequences from robot demonstrations~\citep{chi2023diffusion}, and RT-1 studies how model and data scale affect a multi-task real-robot policy~\citep{brohan2023rt1}. RT-2 co-finetunes VLMs on robot trajectories and web vision-language data with actions represented as tokens~\citep{zitkovich2023rt2}; OpenVLA similarly provides a pretrained, adaptable vision-language-action policy~\citep{kim2025openvla}. Reflective VLA trains an action expert to condition on observation--action--consequence histories, exposing deployment-specific actuation and calibration effects~\citep{lian2026reflective}. Harness VLA instead keeps a VLA frozen as a retryable contact primitive and uses execution memories to delimit the operating range of a fixed primitive library~\citep{zhang2026harness}. \method\ instead controls the robot by interfacing a general VLM through a harness. Our matched evaluation therefore compares harnesses under the same VLM setting rather than trained VLA policies.

\tightpara{Explicit geometry and demonstration-conditioned transfer.}
Task and motion planning couples task-level choices with continuous geometric and motion constraints~\citep{garrett2021tamp}. With foundation models, VoxPoser constructs 3D value maps~\citep{huang2023voxposer}, ReKep generates relational keypoint constraints~\citep{huang2025rekep}, CoPa specifies grasp-part and post-grasp pose constraints~\citep{huang2024copa}, and RoboPoint predicts image-space affordance points~\citep{yuan2025robopoint}. These methods motivate explicit spatial structure for action generation.

Demonstrations provide another route: KAT and RoboPrompt encode examples for in-context action prediction~\citep{dipalo2024kat,yin2025roboprompt}; Vid2Robot learns from paired human videos and robot trajectories~\citep{jain2024vid2robot}; XSkill learns cross-embodiment skill prototypes~\citep{xu2023xskill}. UMI combines hand-held gripper demonstrations, relative action trajectories, and latency matching to train transferable robot policies~\citep{chi2024umi}. Our off-task trajectory instead initializes supported motion correspondence, functional-part evidence, and camera context. It supplies body knowledge without prescribing the new task's sequence or training a visuomotor policy.

\tightpara{Body models and observation.}
Continuous self-modeling enables recovery after morphological damage~\citep{bongard2006resilient}, and self-perception can recover a body scheme without a supplied kinematic model~\citep{sturm2008bodyscheme}. Later models learn forward behavior from action--sensation data~\citep{kwiatkowski2019selfmodel} or pose-conditioned occupancy for planning and damage detection~\citep{chen2022selfmodel}. HumanCLAW studies VLM body reasoning in simulated egocentric tasks, separating action decisions from balance and motor-tracking failures~\citep{li2026humanclaw}. Our body representation is partial: supported motion directions, functional points, temporary tool geometry, and observation context. Multi-view guidance in Show-Harness and iterative visual queries in PIVOT inform observation selection~\citep{chen2026showharness,nasiriany2024pivot}; we also consider how a proposed action affects subsequent visibility.

\tightpara{Experience, memory, and validity under change.}
Inner Monologue feeds observations and success feedback into language-model planning~\citep{huang2023innermonologue}; REFLECT summarizes multisensory robot histories to explain failures and guide correction~\citep{liu2023reflect}. Reflexion retains verbal feedback~\citep{shinn2023reflexion}, and ExpeL retrieves insights extracted from prior tasks~\citep{zhao2024expel}. RoboCat incorporates self-generated trajectories into subsequent training~\citep{bousmalis2024robocat}, whereas PhysMem verifies interaction hypotheses before reuse~\citep{li2026physmem}. ASPIRE diagnoses failures from execution traces, validates repairs through re-execution, and consolidates reusable fixes into a growing skill library~\citep{lu2026aspire}. AGP retains procedures and corrections~\citep{jia2026agentpolicy}, while GPT-Policy adapts through demonstration and interaction context~\citep{cheng2026gptpolicy}. Relational transition models also support planning across object instances and configurations~\citep{chitnis2022nsrt}.

These works establish feedback, experience reuse, and validation as shared foundations. \method\ addresses a specific dependency within this process: correcting body geometry can change the conditions inferred from an unchanged historical interaction. It therefore recomputes those conditions from the original observations and revalidates the affected rules before reuse. The distinction concerns how experience is maintained when its geometric basis changes, rather than the existence of memory or a verification loop.

\begin{figure*}[t]
\centering
\includegraphics[width=\textwidth]{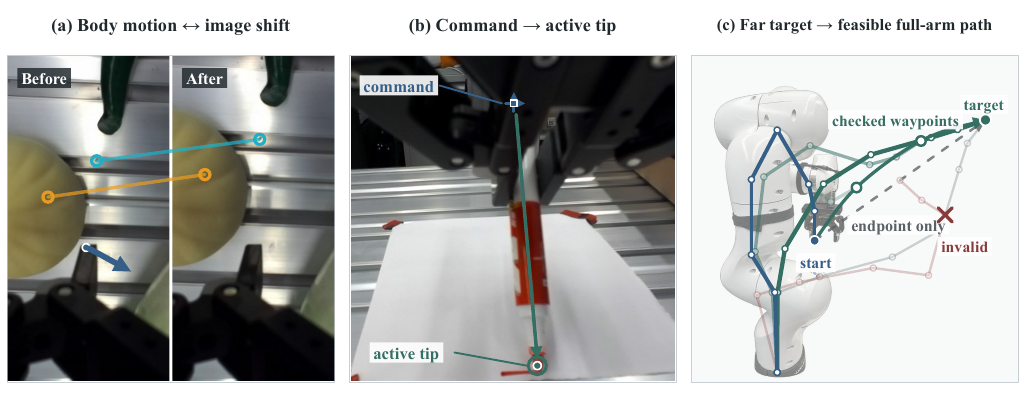}
\caption{\textbf{Action-relevant relations exposed by the body model.} \textbf{(a)} Achieved robot motion is paired with the displacement of stationary landmarks in the wrist image, supporting visual-direction queries within the observed domain. \textbf{(b)} Permanent end-effector geometry is composed with the current attachment to map the commanded frame to the acting tool tip. The blue glyph schematically denotes the declared \texttt{panda\_link8} frame; the green point marks the visible writing tip. \textbf{(c)} For a distant target, a valid endpoint solution does not certify the intervening motion. The translucent cutout anchors the current physical embodiment; the overlaid paths and body poses schematically contrast an endpoint-only transition with a discontinuous intermediate IK branch (red) against sequentially seeded waypoints that preserve a continuous full-arm path (green) to the same target. The query evaluates path feasibility using the kinematic relations in $B$ and the current joint and attachment state in $M$ before execution.}
\label{fig:pipeline}
\end{figure*}

\section{Problem and Learning Protocol}
\label{sec:problem}
At model--tool decision $i$, a  general-purpose VLM receives instruction $\ell$, current images and available depth $o_t$, measured pose $T_t=(R_t,p_t)$, gripper state, and available joint feedback. The index $t$ advances with fresh observations and physical transitions, not with every read-only query. Declared robot and motion limits are supplied as interface constraints. The output $u_i$ is a query, observation request, motion proposal, or termination request. A motion proposal specifies a chunk $A_t$ of numerical Cartesian segments and optional gripper commands; a local executor performs kinematics and timing and records the achieved movement.

One continuous off-task episode initializes the body model:
\begin{equation}
 \tau_0=\{(o_m,a_m^{\rm req},a_m^{\rm ach},o_{m+1})\}_{m=1}^{L},
 \qquad B_0=\mathcal C(\tau_0,\mathcal H),
\end{equation}
Here $\mathcal H$ contains declared hardware metadata. The trajectory seeds supported motion--response relations by pairing observations with requested and achieved actions; independently versioned geometric measurements populate the corresponding functional-geometry components. A body snapshot $B_k$ stores supported motion response, functional geometry, camera bindings, and component versions. Knowledge $K_k$ stores conditional outcomes and their body dependencies; $M_t$ retains temporary episode facts such as tool attachment. The policy interface is
\begin{equation}
 u_i\sim\pi_\theta\!\left(\ell,o_t,\mathcal R(B_k,K_k;o_t),M_t\right),
 \label{eq:policy}
\end{equation}
where $\mathcal R$ retrieves applicable evidence and model weights $\theta$ remain fixed. Adaptation changes $B_k$ and $K_k$, not the VLM. Raw events remain immutable, and a fixed evaluation block cannot use its own outcomes to update its snapshot. Paused continuations of one physical episode remain one statistical group. The fixed-budget comparison and continual sequences in \cref{sec:experiments} evaluate task completion and repeated-task decision effort separately.

\section{Method}
\label{sec:method}
\method\ organizes robot interaction around a queryable, partial body model. Before acting, the model relates motion and functional-part geometry to the commands chosen by a frozen VLM. After acting, it supports checks of the observed response and supplies the geometric conditions under which experience is interpreted. These two roles couple body estimation to experience reuse: correcting a body estimate can change an earlier interaction's conditions without changing its observed outcome. We describe the representation and initialization, body-conditioned interaction, evidence interpretation, and dependency-aware revision.

\subsection{Body Representation and Initialization}
\label{sec:body}
\tightpara{A partial, queryable body model.}
We define body knowledge as action-relevant relations supported by observations and accessible through queries. \Cref{fig:pipeline} previews three complementary forms in the order used for action reasoning: a local visual response relates achieved body motion to wrist-image displacement and supports inverse direction queries; permanent end-effector geometry combines with the current attachment to locate the active tip or part; and full-arm kinematic information distinguishes endpoint reachability from a continuous feasible path. In the last case, $B$ provides the declared chain and its supported kinematic relations, while $M$ provides the measured joint state and current attachment. Their query result conditions planner choices; the executor rechecks the selected motion against fresh state and retains final authority. A versioned body is
\begin{equation}
 B_k=(C_k,G_k,\mathsf R_k,D_k,V_k).
 \label{eq:body_representation}
\end{equation}
Here $C_k$ identifies the physical configuration, declared kinematic chain, and camera acquisition modes; $G_k$ represents permanent functional geometry; $\mathsf R_k$ represents the identifiable local motion--observation response; $D_k$ specifies the supported operating domain; and $V_k$ records sources and validation. A query returns a value with its frame, units, support, and available uncertainty. Relations outside the evidence-supported domain return $\unknown$. The representation is therefore partial: it exposes identifiable relations rather than assuming a complete dynamics model.

\tightpara{Separate body, state, and experience.}
Evidence $\mathcal E$ preserves observations, achieved motion, and attributed outcome records. Working state $M_t$ describes the current scene, progress, and temporary attachments. Knowledge $K_k$ stores validated conditional action--effect relations together with their supporting evidence and body dependencies. These stores have different update rules: new observations revise current state, whereas reusable body estimates and experience require validation.

For a held marker, permanent finger geometry belongs to $G_k$, while the grip-to-tip relation belongs to $M_t^{\mathrm{attach}}$. Both contribute to a functional-part query, but regrasping invalidates the attachment rather than the permanent finger model. Each episode uses a fixed body version $B_{\mathrm{pin}}$ and a compatible knowledge snapshot while updating $M_t$. Contradictory evidence can suspend an affected query or rule during the episode; a replacement is activated only after validation.

\tightpara{Initialize from one off-task trajectory.}
The trajectory supplies observation--motion pairs, not a task-specific action sequence to imitate. For successive measured end-effector poses $(R_m,p_m)$ and $(R_{m+1},p_{m+1})$, the achieved translation in the initial end-effector frame is $d_m^{\mathrm{eef}}=R_m^\top(p_{m+1}-p_m)$. We pair this displacement with its actual before/after observations to initialize identifiable relations in $B_0$. Each component records whether it derives from the seed trajectory, declared hardware, an independent measurement, or a later interaction.

The trajectory initializes a local visual response. For a rigid wrist camera, static scene points, and near-pure translation, normalized image correspondences $\xi,\xi'\in\mathbb R^2$ satisfy
\begin{equation}
 Z'(\xi'-\xi)\approx J_B(\xi)d^{\mathrm{eef}},
 \label{eq:method_projection}
\end{equation}
where $Z'$ is observed after-motion depth and $J_B(\xi)\in\mathbb R^{2\times3}$ is the fitted response. We reject inconsistent correspondences and fit within the translation subspace excited by the trajectory, balancing physical transitions so that dense image tracks do not dominate. This relation supports retrospective response checks and the direction query below; independently sourced measurements supply metric functional geometry. The parameterization and estimation procedure are given in \cref{app:body_estimation}.

\subsection{Body-Conditioned Interaction}
\label{sec:interaction}
\tightpara{Assemble evidence for the next decision.}
At each decision, the harness supplies the VLM with the task goal and completion criteria, current images and measured robot state, working state $M_t$, relevant body-query results, and applicable knowledge. Selected records from $\mathcal E$ provide historical context when needed. Rule applicability is checked before relevance: a rule must match the current configuration, body dependencies, and observed conditions. Undetermined conditions identify evidence to acquire, while unresolved counterevidence remains visible.

The VLM can query a body relation, request an observation, propose motion, or request termination. Tool results enter subsequent model inputs, allowing several queries before a physical action. After execution, the next decision is conditioned on achieved motion and its observations. The VLM weights remain frozen throughout.

\tightpara{Query a motion direction.}
The local visual response connects a desired image direction to robot translation. Let $s\in\mathbb R^2$ be the desired image direction at $\xi$, and let the columns of $U$ form an orthonormal basis for the excited translation subspace. The query solves
\begin{equation}
 d_0=U[J_B(\xi)U]^\dagger s,\qquad
 \hat d^{\mathrm{base}}=R_t\frac{d_0}{\|d_0\|_2},
 \label{eq:motion_query}
\end{equation}
where $\dagger$ denotes the pseudoinverse and $R_t$ maps end-effector coordinates to the base frame. Positive depth scales image displacement without changing its direction, so this query does not require future depth. It returns a direction when support, conditioning, and solution checks pass. The VLM selects travel distance and orientation from the task and current observations; the query supplies a motion relation rather than a task policy.

\tightpara{Convert a functional target into a control target.}
Manipulation often acts through a fingertip, finger surface, or tool edge rather than the end-effector origin. For a functional part $j$ with a supported point model,
\begin{equation}
 x_j=p_t+R_t r_j(g_t),
 \label{eq:functional_point}
\end{equation}
where $(R_t,p_t)$ is the measured end-effector pose, $g_t$ is gripper state, and $r_j$ is the end-effector-frame offset determined by $B_{\mathrm{pin}}$ and the current attachment. Given a target $x_o$ and unit approach direction $n$ in the same base frame, the query computes
\begin{equation}
 \begin{aligned}
 \delta&=x_o-x_j,\qquad h=n^\top\delta,\\
 e_\perp&=\|(I-nn^\top)\delta\|_2.
 \end{aligned}
 \label{eq:relations}
\end{equation}
The axial gap $h$ and lateral offset $e_\perp$ distinguish approaching the target from aligning with it. Edge and surface interactions use their corresponding supported geometry rather than treating a point as the full interaction region.

For a desired part position $x_j^*$, orientation $R^*$, and gripper state $g^*$ selected by the VLM, the control target is
\begin{equation}
 p^*=x_j^*-R^*r_j(g^*).
 \label{eq:functional_inverse}
\end{equation}
This compensates for the offset between the commanded origin and the part producing the effect. Geometric queries require measured offsets and coordinate transforms. When uncertainty prevents a relation from determining a condition, the condition remains unknown and can motivate further observation.

\tightpara{Expose full-arm feasibility.}
A target pose is not a complete motion plan. For a sufficiently long or multi-waypoint proposal, a body query first solves the target pose and then evaluates the intervening route. Within each proposed segment it samples position linearly and orientation by quaternion interpolation, solves each pose from the preceding joint solution, and checks the resulting full-arm path. The query reports endpoint reachability separately from continuous-path feasibility, including the first failed stage when the latter is rejected. It is bound to the pinned kinematic description in $B$ and the measured joint state and attachment revision in $M$; unmeasured scene or attachment clearance remains unknown. Before dispatch, the executor repeats inverse-kinematics, joint-limit, workspace, modeled self-collision, and retiming checks against fresh telemetry, so an earlier query cannot override a later rejection.

\tightpara{Execute, then observe.}
The VLM proposes a chunk of Cartesian waypoints and gripper commands with an observable expected effect. Chunk boundaries follow the need for new evidence: a writing stroke can proceed after contact is established, whereas uncertain first contact calls for observation. The executor applies fixed motion constraints $S$, performs kinematics and timing, and returns execution status, achieved motion, and new observations. Body learning does not modify $S$. These records establish what was executed; the next stage determines its effect.

\subsection{Evidence Interpretation and Episode State}
\label{sec:evidence}
\tightpara{Distinguish execution, body response, and task effect.}
An unsuccessful interaction can originate at different stages. \method\ asks whether the requested motion occurred, whether the observed response agrees with the supported body model, and whether the intended effect appeared. Each judgment uses different evidence: a rejected request concerns execution feasibility; achieved motion with reliable correspondences tests $\mathsf R$; independent geometric measurements inform $G$; and outcomes observed over a specified effect window support conditional knowledge in $K$. Occlusion or unobserved outcomes yield $\unknown$, not \emph{no effect}.

Each interaction therefore preserves before/after observations, requested and achieved motion, execution status, and the source of its outcome assessment. Interpretations can be revised while the original observations and actions remain unchanged. This separation directs investigation toward the relevant stage instead of treating every task failure as evidence against the body model.

\tightpara{Update current state without changing historical outcomes.}
New observations update scene, attachment, and progress estimates in $M_t$. Object motion or regrasping prompts reassessment of relations that depend on the old state. A fresh image can establish the current scene, but does not replace an earlier action's direct after-observation. Likewise, improved visibility provides information, not evidence that the robot changed an object's state. Current observations thus support replanning while preserving the provenance of externally caused scene changes.

\tightpara{Check key outcomes in a separate context.}
Ordinary feedback enters the next planning decision. Grasp/release transitions, proposed milestones, persistent lack of progress or contradictions, and completion requests trigger a separate verification call. The verifier receives fixed criteria, relevant observations, achieved motion, and unresolved counterevidence, without the planner's account of its own success. Its conclusion is scoped to the event being checked: confirming a grasp does not confirm the whole task. Declaring completion requires a passing assessment and resolution of outstanding objections; an inconclusive assessment instead motivates observation or recovery. Because the planner and verifier may share a VLM, reusable body and knowledge updates are governed by the measurement-based validation below.

\tightpara{Acquire evidence to distinguish explanations.}
If a marker leaves no visible ink, plausible causes include absent contact, a changed grip, occlusion, or an ineffective marker. The planner first checks achieved motion and existing observations, then identifies an observation or diagnostic action that distinguishes the remaining explanations. Evidence of a changed grip updates the attachment in $M_t$; a persistent discrepancy supported by qualified measurements motivates a body candidate. Diagnostic actions use the same execution constraints and resource budget as task actions, with a stated observable expectation and stopping condition.

\subsection{Dependency-Aware Revision}
\label{sec:revision}
\tightpara{A body correction changes experience conditions.}
An experience describes not just an action and outcome, but the body relations under which they occurred. Suppose a permanent fingertip offset is corrected by $\Delta r_j$ under the same physical configuration. Holding the recorded pose, target, and approach direction fixed, \cref{eq:relations} gives
\begin{equation}
 \Delta h=-n^\top R_t\Delta r_j.
 \label{eq:gap_revision}
\end{equation}
The observed object motion is unchanged, but the reconstructed fingertip--target gap differs. A rule conditioned on the old gap can therefore assign that interaction to a different condition. Updating the body while reusing every old rule unchanged would mix incompatible interpretations of the same evidence.

\tightpara{Propose and validate body revisions.}
Current-state corrections support immediate replanning; reusable body revisions undergo separate validation. New motion evidence produces a candidate response fit, while permanent geometry requires independent geometric observations in a common frame. Temporary attachment changes remain in $M_t$.

After fitting on discovery data, the candidate is frozen and evaluated on independent interactions. Response validation checks improvement in newly encountered conditions together with accuracy retention in the previously supported domain. Repeated attempts within an episode do not count as independent evidence. Candidates that pass become available after the episode ends, preserving the ongoing episode's body snapshot. Estimators and validation criteria are detailed in \cref{app:body_estimation,app:validation}.

\tightpara{Learn conditional action--effect relations.}
Reusable experience specifies which observed conditions distinguish an action's outcomes. We extract features $\phi(e;B_k)$ from an interaction $e$ using its event-time configuration and attachment. We consider two outcome types: detachment after holding, and no discernible translation of a selected target region after approach. The latter concerns the measured translation, not the absence of contact or rotation. Rules apply within matched action, orientation, support, surface-appearance, and functional-role conditions.

For a scalar relation feature, discovery compares a finite set of threshold splits and estimates the outcome probability in each branch. A candidate must improve prediction over a same-context predictor that omits the feature, then pass independent validation with the rule frozen. Each rule retains its support, counterexamples, evidence, and body dependencies. The resulting rules are scoped conditional predictors. \Cref{app:rule_estimation} specifies the outcome measurements, estimator, and validation objective.

\tightpara{Reinterpret affected experience before reuse.}
Let $\mathcal D_r$ denote the body components used to derive rule $r$, and $\Delta\mathcal D$ the revised components. Revalidation is required when
\begin{equation}
 \mathcal D_r\cap\Delta\mathcal D\ne\varnothing.
 \label{eq:dependency}
\end{equation}
The harness recomputes affected features from the original discovery evidence and refits the rule's numerical conditions while retaining its feature, outcome, and contextual scope. This creates a revised candidate; the affected rule is withheld under the new body until independent validation passes. Evidence insufficient for remeasurement leaves it withheld. Rules independent of the changed components remain eligible for reuse.

Reinterpretation reuses the original observations, outcomes, and issued predictions while updating only their estimated body-dependent conditions. It applies to an estimate correction within a physical configuration. An installation or camera-mode change instead establishes a new configuration, whose geometry is not applied retrospectively to old observations; regrasping changes the temporary attachment. Subsequent episodes reuse validated body knowledge and compatible experience while rebuilding current state from new observations.

\section{Real-Robot Evaluation}
\label{sec:experiments}
We evaluate task completion under fixed reasoning budgets and reasoning effort during continual interaction. The first protocol tests the initialized harness without accumulated experience; the second tracks successful completions as knowledge accumulates from both successful and failed attempts.

\begin{figure}[t]
\centering
\includegraphics[width=\columnwidth]{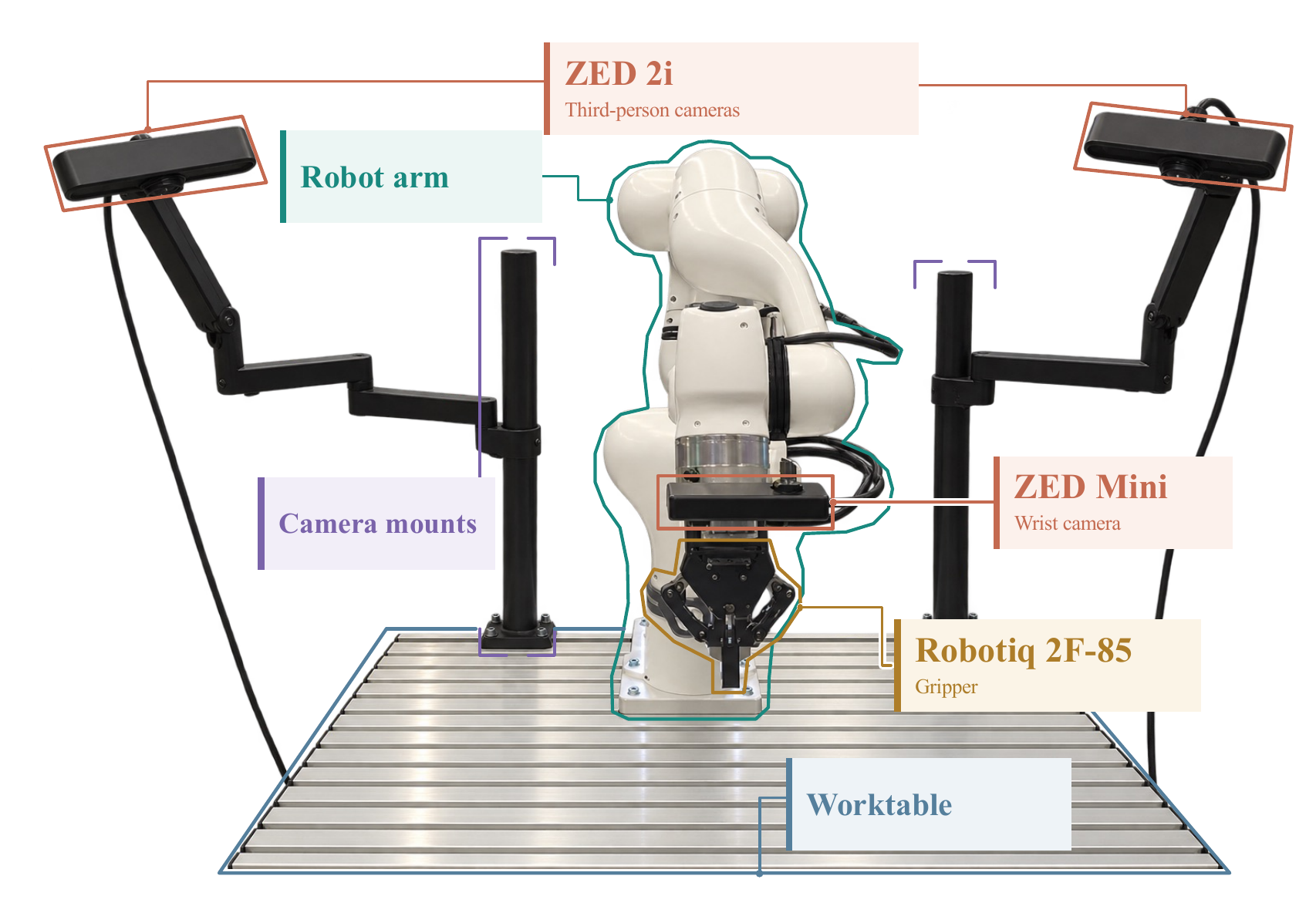}
\caption{\textbf{Real-robot evaluation setup.} A Franka Research 3 with a Robotiq 2F-85 gripper operates over the worktable. A ZED Mini provides the wrist view, while two ZED 2i cameras provide third-person views. The same robot and camera arrangement is used in both evaluation protocols.}
\label{fig:robot_setup}
\end{figure}

\begin{figure*}[t]
\centering
\includegraphics[width=\textwidth]{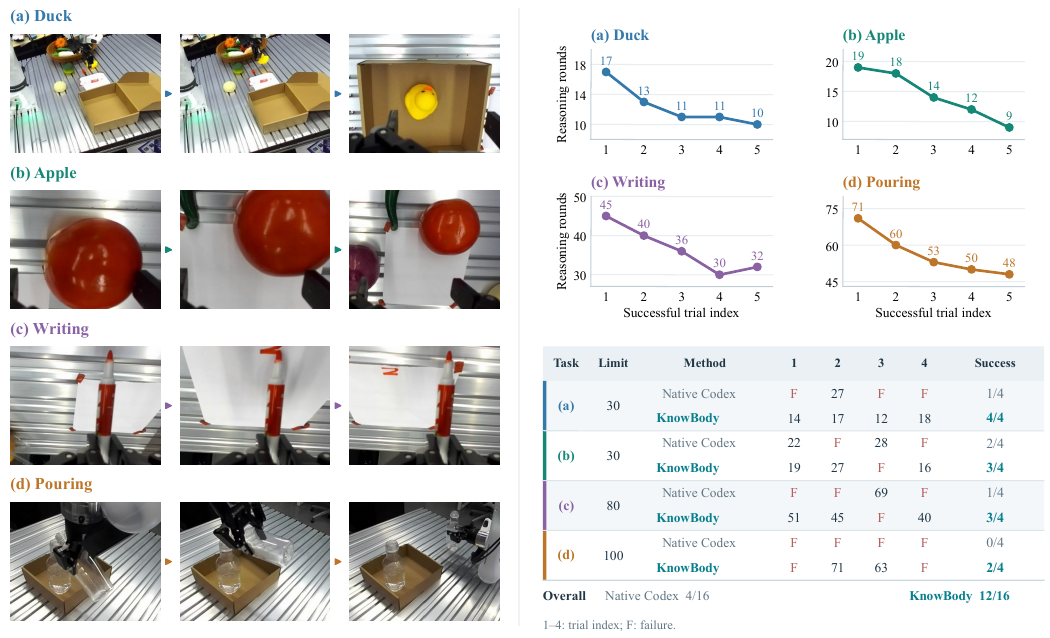}
\caption{\textbf{Task examples and two evaluation protocols.} \textbf{Left:} illustrative development sequences for duck placement, apple pushing, writing, and pouring. \textbf{Upper right:} reasoning rounds for five successive successful completions per task under the continual-interaction protocol. Knowledge is retained across attempts and tasks; intervening failures contribute updates but are not plotted. The horizontal axis indexes successful trials, and vertical ranges differ across tasks.\textbf{Lower right:} all 32 fixed-budget trials, with four trials per method and task. Native Codex denotes the native Codex harness using the same frozen VLM as \method.Numbers denote reasoning rounds for successful trials; F denotes failure. \method\ starts each trial from the same initialization with cross-episode updates disabled. Objects are randomly repositioned before each attempt in both protocols.}
\label{fig:task_results}
\end{figure*}

\subsection{Tasks and metrics}
\label{sec:setup}
As shown in \cref{fig:robot_setup}, we use a Franka Research 3 with a Robotiq 2F-85 gripper. A ZED Mini provides the wrist view, and two ZED 2i cameras provide third-person views of the worktable. Both methods use GPT6 with high reasoning effort and frozen model weights. Four tasks exercise different aspects of body knowledge: (a) place a yellow duck into a paper box, (b) push an apple onto a white-paper region, (c) grasp a marker and write a digit, and (d) pour water from a mineral-water bottle into an empty bottle. The respective reasoning-round limits are 30, 30, 80, and 100.

Task objects are randomly repositioned before each attempt in both protocols. A trial is unsuccessful if it exceeds the reasoning budget or cannot continue autonomously, including after an object drop. We report completion within budget and reasoning rounds for successful trials. A round is a model decision in the interaction loop, rather than a robot motion segment.

\subsection{Fixed-budget comparison}
\label{sec:task_comparison}
\textbf{Direct} uses the base Codex agent loop with its built-in interaction-history management and automatic context compaction~\citep{bolin2026codexloop}. \textbf{KnowBody} uses our harness initialized from one off-task trajectory of grasping bread and placing it in a basket. Continual learning is disabled: each \method\ trial starts from the same initialization and inherits no updates from previous trials. We conduct four trials per method and task, totaling 32 trials. This comparison evaluates the combined reference-and-harness design.

\Cref{fig:task_results} (lower right) reports every outcome. \method\ completes 12/16 trials (75\%), compared with 4/16 (25\%) for Direct. Success increases from 1/4 to 4/4 for duck placement, 2/4 to 3/4 for pushing, 1/4 to 3/4 for writing, and 0/4 to 2/4 for pouring. Thus, the initialized harness improves completion on all four tasks without relying on experience accumulated across these trials.

Among successful trials, mean reasoning counts decrease from 27.0 to 15.3 for duck placement, 25.0 to 20.7 for pushing, and 69.0 to 45.3 for writing, corresponding to reductions of 17--44\%. Because Direct completes no pour trials, the 67.0-round \method\ mean has no successful-baseline counterpart. These averages are conditioned on successful trials.

\textbf{Comparison with DROID-trained policies.}
We additionally evaluate $\pi_{0.5}$-DROID and $\pi_{0}$-DROID on the same robot, task definitions, and success criteria. Both checkpoints are evaluated as provided, without task-specific adaptation. Each policy is tested for 20 trials per task; the \method\ results are from the four fixed-budget trials described above.

\begin{table}[t]
\centering
\small
\setlength{\tabcolsep}{3.5pt}
\caption{\textbf{Task success on the DROID setup.}
Entries report successful trials over total attempts.}
\label{tab:droid_policy_comparison}
\begin{tabular}{lcccc}
\toprule
Method & Duck & Apple & Writing & Pouring \\
\midrule
\method                 & 4/4   & 3/4  & 3/4  & 2/4  \\
$\pi_{0.5}$-DROID       & 19/20 & 2/20 & 0/20 & 0/20 \\
$\pi_{0}$-DROID         & 18/20 & 0/20 & 0/20 & 0/20 \\
\bottomrule
\end{tabular}
\end{table}

As shown in \cref{tab:droid_policy_comparison}, \method\ succeeds across all four tasks. The DROID-trained policies perform strongly or partially on duck placement, while only $\pi_{0.5}$-DROID records additional successes on apple pushing; neither policy completes writing or pouring.

\subsection{Continual interaction}
\label{sec:continual_results}
We then enable persistent updates to body knowledge and experience. Each task is attempted until five successful completions are recorded, after which evaluation switches to the next task. Knowledge is retained both between attempts and across tasks. Failed attempts between recorded successes also contribute updates; only model weights remain frozen. Each new attempt uses a randomized object arrangement.

\Cref{fig:task_results} (upper right) plots reasoning rounds against \emph{successful-trial index}, rather than total attempt count. From the first to fifth recorded success, rounds decrease from 17 to 10 for duck placement, 19 to 9 for pushing, 45 to 32 for writing, and 71 to 48 for pouring: reductions of 41.2\%, 52.6\%, 28.9\%, and 32.4\%, respectively. Writing is non-monotonic, rising from 30 to 32 rounds at the final point.

These trends indicate lower decision effort on later successful completions. We interpret the curves conditional on successful completion; completion over all fixed-budget trials is reported separately above.

\section{Discussion}
\label{sec:discussion}
The fixed-budget comparison evaluates KnowBody as an integrated harness under a matched VLM and reasoning budget. Across four tasks on one robot, with four trials per method--task condition, completion increases from 4/16 to 12/16. Reasoning-round averages are reported over successful trials.

The continual protocol evaluates how decision effort evolves as validated knowledge is retained across interactions and tasks. Its curves index five successive successful completions per task, while the fixed-budget protocol reports every trial outcome. Together, the two protocols separate completion under a fixed initialization from decision effort during continual interaction.

\section{Conclusion}
\method\ organizes direct VLM control around explicit body knowledge initialized from an off-task trajectory and revised through interaction. Functional geometry and informative views connect general task reasoning to numerical robot actions, while dependency-aware revision keeps reusable experience consistent with updated body estimates. Across four real-robot tasks, KnowBody improves completion under fixed reasoning budgets, and later successful completions require fewer reasoning rounds during continual interaction. Together, these results support explicit, revisable body knowledge as an interface between frozen VLM reasoning, numerical action, and experience reuse.

{\fontsize{9}{10}\selectfont\bibliographystyle{plainnat}\bibliography{references}}

@inproceedings{yin2025roboprompt,
  title={In-Context Learning Enables Robot Action Prediction in {LLMs}},
  author={Yin, Yida and Wang, Zekai and Sharma, Yuvan and Niu, Dantong and Darrell, Trevor and Herzig, Roei},
  booktitle={IEEE International Conference on Robotics and Automation (ICRA)},
  pages={6117--6124},
  year={2025},
  url={https://arxiv.org/abs/2410.12782}
}

@inproceedings{dipalo2024kat,
  title={Keypoint Action Tokens Enable In-Context Imitation Learning in Robotics},
  author={Di Palo, Norman and Johns, Edward},
  booktitle={Proceedings of Robotics: Science and Systems},
  year={2024},
  doi={10.15607/RSS.2024.XX.096},
  url={https://www.roboticsproceedings.org/rss20/p096.html}
}

@inproceedings{nasiriany2024pivot,
  title={{PIVOT}: Iterative Visual Prompting Elicits Actionable Knowledge for {VLMs}},
  author={Nasiriany, Soroush and Xia, Fei and Yu, Wenhao and Xiao, Ted and Liang, Jacky and Dasgupta, Ishita and Xie, Annie and Driess, Danny and Wahid, Ayzaan and Xu, Zhuo and Vuong, Quan and Zhang, Tingnan and Lee, Tsang-Wei Edward and Lee, Kuang-Huei and Xu, Peng and Kirmani, Sean and Zhu, Yuke and Zeng, Andy and Hausman, Karol and Heess, Nicolas and Finn, Chelsea and Levine, Sergey and Ichter, Brian},
  booktitle={Proceedings of the 41st International Conference on Machine Learning},
  pages={37321--37341},
  volume={235},
  series={Proceedings of Machine Learning Research},
  year={2024},
  url={https://proceedings.mlr.press/v235/nasiriany24a.html}
}

@inproceedings{huang2025rekep,
  title={{ReKep}: Spatio-Temporal Reasoning of Relational Keypoint Constraints for Robotic Manipulation},
  author={Huang, Wenlong and Wang, Chen and Li, Yunzhu and Zhang, Ruohan and Fei-Fei, Li},
  booktitle={Proceedings of the 8th Conference on Robot Learning},
  volume={270},
  series={Proceedings of Machine Learning Research},
  pages={4573--4602},
  year={2025},
  note={Conference held in 2024; proceedings published in 2025},
  url={https://proceedings.mlr.press/v270/huang25g.html}
}

@inproceedings{huang2023voxposer,
  title={{VoxPoser}: Composable {3D} Value Maps for Robotic Manipulation with Language Models},
  author={Huang, Wenlong and Wang, Chen and Zhang, Ruohan and Li, Yunzhu and Wu, Jiajun and Fei-Fei, Li},
  booktitle={Proceedings of the 7th Conference on Robot Learning},
  volume={229},
  series={Proceedings of Machine Learning Research},
  pages={540--562},
  year={2023},
  url={https://proceedings.mlr.press/v229/huang23b.html}
}

@inproceedings{liang2023code,
  title={Code as Policies: Language Model Programs for Embodied Control},
  author={Liang, Jacky and Huang, Wenlong and Xia, Fei and Xu, Peng and Hausman, Karol and Ichter, Brian and Florence, Pete and Zeng, Andy},
  booktitle={IEEE International Conference on Robotics and Automation (ICRA)},
  pages={9493--9500},
  year={2023},
  doi={10.1109/ICRA48891.2023.10160591},
  url={https://arxiv.org/abs/2209.07753}
}

@misc{zhang2026harness,
  title={Harness {VLA}: Steering Frozen {VLAs} into Reliable Manipulation Primitives via Memory-Guided Agents},
  author={Zhang, Yixian and Zhang, Huanming and Gao, Feng and Li, Xiao and Liu, Zhihao and Zhu, Chunyang and Qiu, Jiaxing and Yan, Yuchen and Liu, Jiyuan and Tang, Wenhao and Fang, Zhengru and Nie, Yi and Wei, Changxu and Wang, Yu and Ding, Wenbo and Yu, Chao},
  year={2026},
  howpublished={arXiv:2607.08448v4},
  url={https://arxiv.org/abs/2607.08448v4}
}

@misc{lian2026reflective,
  title={Reflective {VLA}: In-Context Action Consequences Make {VLAs} Generalize},
  author={Lian, Qing and Yu, Kent and Zhang, Lei},
  year={2026},
  howpublished={arXiv:2606.25215v1},
  url={https://arxiv.org/abs/2606.25215v1}
}

@inproceedings{kim2025openvla,
  title={{OpenVLA}: An Open-Source Vision-Language-Action Model},
  author={Kim, Moo Jin and Pertsch, Karl and Karamcheti, Siddharth and Xiao, Ted and Balakrishna, Ashwin and Nair, Suraj and Rafailov, Rafael and Foster, Ethan P. and Sanketi, Pannag R. and Vuong, Quan and Kollar, Thomas and Burchfiel, Benjamin and Tedrake, Russ and Sadigh, Dorsa and Levine, Sergey and Liang, Percy and Finn, Chelsea},
  booktitle={Proceedings of the 8th Conference on Robot Learning},
  volume={270},
  series={Proceedings of Machine Learning Research},
  pages={2679--2713},
  year={2025},
  note={Conference held in 2024; proceedings published in 2025},
  url={https://proceedings.mlr.press/v270/kim25c.html}
}

@inproceedings{chi2023diffusion,
  title={Diffusion Policy: Visuomotor Policy Learning via Action Diffusion},
  author={Chi, Cheng and Feng, Siyuan and Du, Yilun and Xu, Zhenjia and Cousineau, Eric and Burchfiel, Benjamin and Song, Shuran},
  booktitle={Proceedings of Robotics: Science and Systems},
  year={2023},
  doi={10.15607/RSS.2023.XIX.026},
  url={https://arxiv.org/abs/2303.04137}
}

@inproceedings{shinn2023reflexion,
  title={Reflexion: Language Agents with Verbal Reinforcement Learning},
  author={Shinn, Noah and Cassano, Federico and Gopinath, Ashwin and Narasimhan, Karthik and Yao, Shunyu},
  booktitle={Advances in Neural Information Processing Systems},
  volume={36},
  year={2023},
  doi={10.52202/075280-0377}
}

@article{zhao2024expel,
  title={{ExpeL}: {LLM} Agents Are Experiential Learners},
  author={Zhao, Andrew and Huang, Daniel and Xu, Quentin and Lin, Matthieu and Liu, Yong-Jin and Huang, Gao},
  journal={Proceedings of the AAAI Conference on Artificial Intelligence},
  volume={38},
  number={17},
  pages={19632--19642},
  year={2024},
  doi={10.1609/aaai.v38i17.29936},
  url={https://ojs.aaai.org/index.php/AAAI/article/view/29936}
}

@misc{li2026physmem,
  title={{PhysMem}: Scaling Test-Time Memory for Embodied Physical Reasoning},
  author={Li, Haoyang and You, Yang and Su, Hao and Guibas, Leonidas},
  year={2026},
  howpublished={arXiv:2602.20323v6},
  url={https://arxiv.org/abs/2602.20323v6}
}

@inproceedings{chitnis2022nsrt,
  title={Learning Neuro-Symbolic Relational Transition Models for Bilevel Planning},
  author={Chitnis, Rohan and Silver, Tom and Tenenbaum, Joshua B. and Lozano-P{\'e}rez, Tom{\'a}s and Kaelbling, Leslie Pack},
  booktitle={IEEE/RSJ International Conference on Intelligent Robots and Systems (IROS)},
  pages={4166--4173},
  year={2022},
  doi={10.1109/IROS47612.2022.9981440},
  url={https://arxiv.org/abs/2105.14074}
}

@article{kwiatkowski2019selfmodel,
  title={Task-Agnostic Self-Modeling Machines},
  author={Kwiatkowski, Robert and Lipson, Hod},
  journal={Science Robotics},
  volume={4},
  number={26},
  pages={eaau9354},
  year={2019},
  doi={10.1126/scirobotics.aau9354},
  url={https://doi.org/10.1126/scirobotics.aau9354}
}

@article{chen2022selfmodel,
  title={Full-Body Visual Self-Modeling of Robot Morphologies},
  author={Chen, Boyuan and Kwiatkowski, Robert and Vondrick, Carl and Lipson, Hod},
  journal={Science Robotics},
  volume={7},
  number={68},
  pages={eabn1944},
  year={2022},
  doi={10.1126/scirobotics.abn1944},
  url={https://doi.org/10.1126/scirobotics.abn1944}
}

@misc{chen2026showharness,
  title={{Show-Harness}: Just a {VLM} Agent Can Play Robots},
  author={Chen, Yanzhe and Bai, Zechen and Cao, Zhijun and Zeng, Wenzheng and Lin, Kevin Qinghong and Lin, Yiqi and Liang, Guoqiang and Ma, Kevin Yuchen and Huang, Qiming and Shou, Mike Zheng},
  year={2026},
  howpublished={arXiv:2609.10522v1},
  url={https://arxiv.org/abs/2609.10522v1}
}

@inproceedings{ahn2023saycan,
  title={Do As I Can, Not As I Say: Grounding Language in Robotic Affordances},
  author={Ahn, Michael and others},
  booktitle={Proceedings of the 6th Conference on Robot Learning},
  volume={205},
  series={Proceedings of Machine Learning Research},
  pages={287--318},
  year={2023},
  note={Conference held in 2022; proceedings published in 2023},
  url={https://proceedings.mlr.press/v205/ichter23a.html}
}

@inproceedings{driess2023palme,
  title={{PaLM-E}: An Embodied Multimodal Language Model},
  author={Driess, Danny and Xia, Fei and Sajjadi, Mehdi S. M. and Lynch, Corey and Chowdhery, Aakanksha and Ichter, Brian and Wahid, Ayzaan and Tompson, Jonathan and Vuong, Quan and Yu, Tianhe and Huang, Wenlong and Chebotar, Yevgen and Sermanet, Pierre and Duckworth, Daniel and Levine, Sergey and Vanhoucke, Vincent and Hausman, Karol and Toussaint, Marc and Greff, Klaus and Zeng, Andy and Mordatch, Igor and Florence, Pete},
  booktitle={Proceedings of the 40th International Conference on Machine Learning},
  volume={202},
  series={Proceedings of Machine Learning Research},
  pages={8469--8488},
  year={2023},
  url={https://proceedings.mlr.press/v202/driess23a.html}
}

@inproceedings{brohan2023rt1,
  title={{RT-1}: Robotics Transformer for Real-World Control at Scale},
  author={Brohan, Anthony and others},
  booktitle={Proceedings of Robotics: Science and Systems},
  year={2023},
  doi={10.15607/RSS.2023.XIX.025},
  url={https://www.roboticsproceedings.org/rss19/p025.html}
}

@inproceedings{zitkovich2023rt2,
  title={{RT-2}: Vision-Language-Action Models Transfer Web Knowledge to Robotic Control},
  author={Zitkovich, Brianna and others},
  booktitle={Proceedings of the 7th Conference on Robot Learning},
  volume={229},
  series={Proceedings of Machine Learning Research},
  pages={2165--2183},
  year={2023},
  url={https://proceedings.mlr.press/v229/zitkovich23a.html}
}

@inproceedings{huang2023innermonologue,
  title={Inner Monologue: Embodied Reasoning through Planning with Language Models},
  author={Huang, Wenlong and Xia, Fei and Xiao, Ted and Chan, Harris and Liang, Jacky and Florence, Pete and Zeng, Andy and Tompson, Jonathan and Mordatch, Igor and Chebotar, Yevgen and Sermanet, Pierre and Jackson, Tomas and Brown, Noah and Luu, Linda and Levine, Sergey and Hausman, Karol and Ichter, Brian},
  booktitle={Proceedings of the 6th Conference on Robot Learning},
  volume={205},
  series={Proceedings of Machine Learning Research},
  pages={1769--1782},
  year={2023},
  note={Conference held in 2022; proceedings published in 2023},
  url={https://proceedings.mlr.press/v205/huang23c.html}
}

@inproceedings{zhao2024vlmpc,
  title={{VLMPC}: Vision-Language Model Predictive Control for Robotic Manipulation},
  author={Zhao, Wentao and Chen, Jiaming and Meng, Ziyu and Mao, Donghui and Song, Ran and Zhang, Wei},
  booktitle={Proceedings of Robotics: Science and Systems},
  year={2024},
  doi={10.15607/RSS.2024.XX.106},
  url={https://www.roboticsproceedings.org/rss20/p106.html}
}

@article{bousmalis2024robocat,
  title={{RoboCat}: A Self-Improving Generalist Agent for Robotic Manipulation},
  author={Bousmalis, Konstantinos and others},
  journal={Transactions on Machine Learning Research},
  year={2024},
  url={https://openreview.net/forum?id=vsCpILiWHu}
}

@inproceedings{jain2024vid2robot,
  title={{Vid2Robot}: End-to-end Video-conditioned Policy Learning with Cross-Attention Transformers},
  author={Jain, Vidhi and Attarian, Maria and Joshi, Nikhil J. and Wahid, Ayzaan and Driess, Danny and Vuong, Quan and Sanketi, Pannag R. and Sermanet, Pierre and Welker, Stefan and Chan, Christine and Gilitschenski, Igor and Bisk, Yonatan and Dwibedi, Debidatta},
  booktitle={Proceedings of Robotics: Science and Systems},
  year={2024},
  doi={10.15607/RSS.2024.XX.052},
  url={https://www.roboticsproceedings.org/rss20/p052.html}
}

@inproceedings{xu2023xskill,
  title={{XSkill}: Cross Embodiment Skill Discovery},
  author={Xu, Mengda and Xu, Zhenjia and Chi, Cheng and Veloso, Manuela and Song, Shuran},
  booktitle={Proceedings of the 7th Conference on Robot Learning},
  volume={229},
  series={Proceedings of Machine Learning Research},
  pages={3536--3555},
  year={2023},
  url={https://proceedings.mlr.press/v229/xu23a.html}
}

@inproceedings{huang2024copa,
  title={{CoPa}: General Robotic Manipulation through Spatial Constraints of Parts with Foundation Models},
  author={Huang, Haoxu and Lin, Fanqi and Hu, Yingdong and Wang, Shengjie and Gao, Yang},
  booktitle={IEEE/RSJ International Conference on Intelligent Robots and Systems (IROS)},
  pages={9488--9495},
  year={2024},
  doi={10.1109/IROS58592.2024.10801352},
  url={https://arxiv.org/abs/2403.08248}
}

@inproceedings{yuan2025robopoint,
  title={{RoboPoint}: A Vision-Language Model for Spatial Affordance Prediction in Robotics},
  author={Yuan, Wentao and Duan, Jiafei and Blukis, Valts and Pumacay, Wilbert and Krishna, Ranjay and Murali, Adithyavairavan and Mousavian, Arsalan and Fox, Dieter},
  booktitle={Proceedings of the 8th Conference on Robot Learning},
  volume={270},
  series={Proceedings of Machine Learning Research},
  pages={4005--4020},
  year={2025},
  note={Conference held in 2024; proceedings published in 2025},
  url={https://proceedings.mlr.press/v270/yuan25c.html}
}

@article{bongard2006resilient,
  title={Resilient Machines through Continuous Self-Modeling},
  author={Bongard, Josh and Zykov, Victor and Lipson, Hod},
  journal={Science},
  volume={314},
  number={5802},
  pages={1118--1121},
  year={2006},
  doi={10.1126/science.1133687},
  url={https://doi.org/10.1126/science.1133687}
}

@inproceedings{sturm2008bodyscheme,
  title={Unsupervised Body Scheme Learning through Self-Perception},
  author={Sturm, J{\"u}rgen and Plagemann, Christian and Burgard, Wolfram},
  booktitle={IEEE International Conference on Robotics and Automation (ICRA)},
  pages={3328--3333},
  year={2008},
  url={https://cvg.cit.tum.de/members/sturmju/research/bodyschema}
}

@misc{jia2026agentpolicy,
  title={Agent as Policy for Robotic Manipulation},
  author={Jia, Mengzhao and Lin, Yang and Zhang, Xixin and Zhang, Zhihan and Liu, Xiaobai and Jiang, Meng},
  year={2026},
  howpublished={arXiv:2609.12541v2},
  url={https://arxiv.org/abs/2609.12541v2}
}

@misc{cheng2026gptpolicy,
  title={In-Context Robot Learning with {VLM} Agents},
  author={Cheng, Dongzhou and Yi, Taoran and Fang, Ye and Zhang, Xingwu and Feng, Fan and Li, Yixuan and Zhuang, Gengxiong and Wang, Rongze and Yang, Shuai and Song, Wei and Xue, Weizhi and Wu, Minyan and Gui, Jie and Wang, Jiaqi and Wu, Tong},
  year={2026},
  howpublished={Technical report},
  url={https://cheng-haha.github.io/GPT-Policy/paper.pdf?v=20260915-repository-rename}
}

@inproceedings{chi2024umi,
  title={Universal Manipulation Interface: In-The-Wild Robot Teaching Without In-The-Wild Robots},
  author={Chi, Cheng and Xu, Zhenjia and Pan, Chuer and Cousineau, Eric and Burchfiel, Benjamin and Feng, Siyuan and Tedrake, Russ and Song, Shuran},
  booktitle={Proceedings of Robotics: Science and Systems},
  year={2024},
  doi={10.15607/RSS.2024.XX.045},
  url={https://www.roboticsproceedings.org/rss20/p045.html}
}

@article{garrett2021tamp,
  title={Integrated Task and Motion Planning},
  author={Garrett, Caelan Reed and Chitnis, Rohan and Holladay, Rachel and Kim, Beomjoon and Silver, Tom and Kaelbling, Leslie Pack and Lozano-P{\'e}rez, Tom{\'a}s},
  journal={Annual Review of Control, Robotics, and Autonomous Systems},
  volume={4},
  pages={265--293},
  year={2021},
  doi={10.1146/annurev-control-091420-084139},
  url={https://doi.org/10.1146/annurev-control-091420-084139}
}

@inproceedings{liu2023reflect,
  title={{REFLECT}: Summarizing Robot Experiences for Failure Explanation and Correction},
  author={Liu, Zeyi and Bahety, Arpit and Song, Shuran},
  booktitle={Proceedings of the 7th Conference on Robot Learning},
  volume={229},
  series={Proceedings of Machine Learning Research},
  pages={3468--3484},
  year={2023},
  url={https://proceedings.mlr.press/v229/liu23g.html}
}

@misc{lu2026aspire,
  title={{ASPIRE}: Agentic /Skills Discovery for Robotics},
  author={Lu, Runyu and Wu, Yubo and Kou, Ethan and Fu, Letian and Xiao, Wenli and Mandlekar, Ajay and Xu, Yinzhen and Shi, Guanya and Goldberg, Ken and Chen, Ang and Chowdhury, Mosharaf and Zhu, Yuke and Fan, Linxi and Wang, Guanzhi},
  year={2026},
  howpublished={arXiv:2607.00272v1},
  url={https://arxiv.org/abs/2607.00272v1}
}

@misc{fu2026capx,
  title={{CaP-X}: A Framework for Benchmarking and Improving Coding Agents for Robot Manipulation},
  author={Fu, Letian and Yu, Justin and El-Refai, Karim and Kou, Ethan and Xue, Haoru and Huang, Huang and Xiao, Wenli and Wang, Guanzhi and Niu, Dantong and Li, Fei-Fei and Shi, Guanya and Wu, Jiajun and Sastry, Shankar and Zhu, Yuke and Goldberg, Ken and Fan, Linxi},
  year={2026},
  howpublished={arXiv:2603.22435v2},
  url={https://arxiv.org/abs/2603.22435v2}
}

@misc{li2026humanclaw,
  title={{HumanCLAW}: Can Vision-Language Models Act Through a Body?},
  author={Li, Siyao and Gu, Jiawei and Liu, Shuai and Hu, Kairui and Li, Zekun and Li, Linjie and Tang, Chengcheng and Wu, Po-Chen and Shugurov, Ivan and Ma, Lingni and Zollhoefer, Michael and An, Sizhe and Mittal, Abhay and Zhao, Amy and Krishna, Ranjay and Li, Manling and Liu, Ziwei and Guo, Chuan},
  year={2026},
  howpublished={arXiv:2607.27180v2},
  url={https://arxiv.org/abs/2607.27180v2}
}

@misc{bolin2026codexloop,
  author       = {Bolin, Michael},
  title        = {Unrolling the {Codex} Agent Loop},
  howpublished = {OpenAI Engineering},
  year         = {2026},
  month        = jan,
  url          = {https://openai.com/index/unrolling-the-codex-agent-loop/}
}
\clearpage
\appendix
\setupappendixnumbering
\twocolumn
\section{Harness Execution and Timescales}
\label{app:harness_execution}

This appendix expands the execution contract summarized in \cref{fig:harness}. The central separation is between a fast interaction loop, which updates current state from every observation, and a slower publication loop, which changes reusable body and interaction knowledge only after validation. The VLM selects task actions, but does not write active body parameters, publish rules, or bypass the executor.

\subsection{Four Clocks}
\label{app:clocks}

KnowBody distinguishes four indices because counting every model or tool exchange as a physical trial would conflate reasoning, sensing, and evidence collection.

\begin{table}[H]
\centering
\small
\setlength{\tabcolsep}{4pt}
\caption{The four clocks used by the harness. A read-only query can occur within a model call without advancing the physical-interaction clock.}
\label{tab:clocks}
\begin{tabularx}{\columnwidth}{@{}lY@{}}
\toprule
\textbf{Clock} & \textbf{Advances when} \\
\midrule
Call $i$ & A model role is invoked. Planner, verifier, and optional evidence-assistant calls are attributed separately. \\
Interaction $t$ & A new observation is captured or a requested action is dispatched and its achieved prefix is observed. Read-only queries do not advance $t$. \\
Episode $e$ & A new physical evidence group begins. A process restart or pause does not by itself create an independent group. \\
Task $\tau$ & A new task contract fixes the goal, success criteria, diagnostic scope, and resource accounting shared by its episodes. \\
\bottomrule
\end{tabularx}
\end{table}

An episode pins $(B_e,K_e)$ at its start. Its working state $M_t$ changes as the scene, attachment, and task progress change, while original events append to $\mathcal E$. A contradicted body result or rule can be withheld immediately, but a replacement version is published only at an episode boundary. This prevents one episode from changing the coordinate system used to interpret its own earlier actions.

\subsection{Component Authority}
\label{app:authority}
The online loop below makes the authority boundaries operational; \cref{tab:authority} then summarizes what each component can produce and what remains outside its authority.

\subsection{One Online Interaction Loop}
\label{app:online_loop}

\Cref{alg:online_loop} makes the requested--planned--achieved distinction explicit. Read-only body and evidence calls are resolved against the same frozen input snapshot before the planner returns a final choice. An observation choice captures a later stationary view; a changed viewpoint requires a motion choice and therefore the same execution checks as any other movement.

\begin{algorithm}[H]
\small
\caption{KnowBody online interaction loop}
\label{alg:online_loop}
\begin{algorithmic}[1]
\Require $T_\tau$, $(B_e,K_e)$, $o_t$, $M_t$, and $\mathcal E$
\While{the task is active}
  \State $c_i \gets \Call{Compile}{T_\tau,o_t,M_t,B_e,K_e,\mathcal E}$
  \State $u_i \gets \Call{FrozenVLM}{c_i}$ with read-only calls bound to $c_i$
  \If{$u_i$ requests observation}
    \State $o_{t+1}\gets\Call{Observe}{}$
    \State $\mathcal E\gets\Call{Append}{\mathcal E,o_{t+1}}$
    \State $M_{t+1}\gets\Call{GroundStateUpdate}{M_t,o_{t+1}}$
  \ElsIf{$u_i$ proposes motion $A_t^{\mathrm{req}}$}
    \State $A_t^{\mathrm{plan}}\gets\Call{Preflight}{A_t^{\mathrm{req}},S,\text{fresh telemetry}}$
    \State $A_t^{\mathrm{ach}},o_{t+1}\gets\Call{ExecuteObserve}{A_t^{\mathrm{plan}}}$
    \State append request, plan, achieved prefix, status, and views to $\mathcal E$
    \State $M_{t+1}\gets\Call{InterpretCurrentState}{M_t,\mathcal E,B_e}$
  \ElsIf{$u_i$ requests completion}
    \State $v\gets\Call{Verify}{T_\tau,o_t,\text{actual receipts},\text{open objections}}$
    \If{$v=\textsc{Pass}$} \State \Return completed task \Else \State append $v$ to $\mathcal E$ \EndIf
  \Else
    \State \Return reported blocker and supporting evidence
  \EndIf
  \State $t\gets t+1$ when a new physical observation exists; $i\gets i+1$
\EndWhile
\end{algorithmic}
\end{algorithm}

The executor records rejection before physical interpretation: a rejected request provides feasibility evidence but no object-effect sample. For an executed request, body-response residuals are conditioned on achieved motion rather than requested motion. Task effects are then assessed from the appropriate after-observation and effect window. These three judgments use different evidence and thresholds.

\begin{table*}[t]
\centering
\small
\setlength{\tabcolsep}{5pt}
\caption{Authority boundaries in KnowBody. Model assessments can propose state changes or candidates, whereas execution and publication pass through separate contracts.}
\label{tab:authority}
\begin{tabularx}{\textwidth}{@{}L{.16\textwidth}L{.23\textwidth}L{.25\textwidth}Y@{}}
\toprule
\textbf{Component} & \textbf{Reads} & \textbf{Produces} & \textbf{Does not authorize} \\
\midrule
Context compiler & Task contract, current observation, $M_t$, pinned $B_e/K_e$, indexed $\mathcal E$ & One bounded, source-linked model input & Actions, body revisions, or rule publication \\
Frozen VLM planner & Compiled text, attached images, read-only tool results & One structured choice and proposed state updates & Hardware execution or writes to active $B/K$ \\
Body-query layer & Frozen observation, body pin, measured state, attachment revision & Supported, unknown, or invalid numerical relation with provenance & Safety or execution \\
Executor and fixed gates $S$ & Requested motion and fresh telemetry & Planned and achieved motion, rejection/interruption status & Task success or reusable knowledge \\
Evidence verifier & Fixed criteria, observations, actual receipts, unresolved counterevidence & Scoped \textsc{Pass}, \textsc{Needs\_Work}, or \textsc{Blocked} review & Motion, changes to criteria, or publication of $B/K$ \\
Revision pipeline & Qualified original evidence, frozen candidates and validation plans & Published body bundle and eligible rules for a later episode & Changes to the current episode's pin \\
\bottomrule
\end{tabularx}
\end{table*}
\FloatBarrier

\section{Representations and Model-Facing Contracts}
\label{app:contracts}

The symbols $\mathcal E$, $B$, $M$, and $K$ denote different truth and update contracts, not four prose summaries. Their serialized forms can all contain text, numbers, and image references; what distinguishes them is who may write them, how long they remain current, and what must be checked before reuse.

\subsection{Stores and Lifetimes}

\begin{table*}[t]
\centering
\small
\setlength{\tabcolsep}{4pt}
\caption{Representation contract. ``Model view'' is the selected projection used for a decision, not necessarily the complete stored object.}
\label{tab:stores}
\begin{tabularx}{\textwidth}{@{}L{.045\textwidth}L{.19\textwidth}L{.19\textwidth}L{.26\textwidth}Y@{}}
\toprule
& \textbf{Authoritative content} & \textbf{Writer and lifetime} & \textbf{Model view} & \textbf{Invalidation or withholding} \\
\midrule
$\mathcal E$ & Immutable observations, requested/planned/achieved motion, receipts, outcomes, reviews, and source hashes & Runtime and tools append events; records persist across episodes & Current images and short causal facts; older records retrieved by stable evidence ID & Never rewritten. A later interpretation appends a new record and cites the original. \\
$B$ & Versioned configuration $C$, permanent geometry $G$, local response $\mathsf R$, support domain $D$, and provenance/validation $V$ & Estimators propose; validation and publication activate a bundle for later episodes & Pin and component versions remain resident; numerical relations enter through supported query results & Configuration mismatch, unsupported domain, contradicted component, or unpublished candidate \\
$M$ & Current scene relations, attachment revision, task progress, hypotheses, counterevidence, and pending tests & Sensor-grounded updates and source-checked model proposals; rebuilt for each episode & Current compact state plus explicit changes and unresolved risks & Entity, camera, gripper, body, or attachment changes stale only dependent claims \\
$K$ & Conditional action--effect rules, scope, outcome, counterexamples, sources, and body dependencies & Learning pipeline publishes after independent validation; persists across compatible episodes & Conditions are checked first; only eligible rules are ranked and shown & False/unknown condition, changed dependency, incompatible configuration, or counterevidence \\
\bottomrule
\end{tabularx}
\end{table*}

All records use stable IDs and source links. Numerical claims additionally carry coordinate frame, unit, time, uncertainty or an explicit unknown value. A numerical program output becomes body knowledge only through the corresponding measurement and validation contract.

\subsection{Task, Evidence, and State Records}

The task contract makes the goal and stopping conditions durable across process continuations. Count limits may be finite or explicitly null while accounting remains active.

\begin{lstlisting}[style=appendixblock,caption={Compact model-facing task and evidence records.},label={lst:task_evidence}]
{
  "task_id": string,
  "task": string,
  "success_criteria": [string, ...],
  "diagnostic_scope": string,
  "max_diagnostic_actions": integer | null,
  "max_analysis_calls": integer | null,
  "max_model_calls": integer | null,
  "stagnation_window": integer
}

{
  "evidence_id": string,
  "kind": "observation" | "outcome" | "review" | "claim" | ...,
  "value": object,
  "observation_id": string | null,
  "source_ids": [string, ...],
  "applicability": string
}
\end{lstlisting}

The current-state projection separates scene relations from causal effects. A scene claim names its dependent entities, views, attachment dependence, and epistemic status. An interaction assessment can cite a causal outcome only when its after-observation is the current observation. Thus a fresh image may update what is visible without claiming that the robot produced the change.

\begin{lstlisting}[style=appendixblock,caption={Compact current-state and validated-rule records.},label={lst:state_rule}]
{
  "scene_claim": {
    "claim_id": string,
    "subject_refs": [string, ...],
    "relation": string,
    "value": string,
    "epistemic_status": "model_assessed" | "hypothesis" | "unknown",
    "views": [camera_role, ...],
    "depends_on_attachment": boolean
  },
  "attachment": {
    "attachment_id": string,
    "revision": string,
    "status": "attached" | "released" | "uncertain",
    "functional_part": string,
    "source_observation_id": string
  }
}

{
  "rule_id": string,
  "condition": object,
  "action_relation": string,
  "effect": string,
  "support": object,
  "counterexample_ids": [string, ...],
  "source_event_ids": [string, ...],
  "body_dependencies": [component_version, ...],
  "validation_report_id": string
}
\end{lstlisting}

\FloatBarrier

\subsection{Body Queries}

The model does not receive the estimator internals or the entire body ledger. It asks for an action-relevant relation bound to the current observation and episode pin. The three principal query families are: (i) \texttt{image\_direction}, which maps a desired wrist-image direction to a unit translation direction within the excited response subspace; (ii) \texttt{active\_part\_relation}, which composes permanent geometry with the current attachment and target measurement; and (iii) \texttt{motion\_feasibility}, which evaluates endpoint and sampled-path feasibility using the pinned chain, measured joints, and attachment revision. A depth measurement tool can supply camera-frame geometry without converting it into a body component by itself.

\begin{lstlisting}[style=appendixblock,caption={Common body-query response.},label={lst:body_query}]
{
  "status": "supported" | "unknown" | "invalid",
  "query_type": string,
  "observation_id": string,
  "body_id": string,
  "bundle_id": string,
  "result": object | null,
  "frame": string,
  "unit": string,
  "error": object | "unknown",
  "component_versions": object,
  "source_ids": [string, ...],
  "support": object,
  "limitations": [string, ...],
  "executed": false,
  "safety_checked": false
}
\end{lstlisting}

\texttt{supported} means that the requested relation is available within its recorded domain. \texttt{unknown} identifies absent measurement, coverage, or compatible configuration. \texttt{invalid} identifies a malformed or stale request. Even a supported result is read-only: fresh telemetry and fixed gates are checked again before dispatch.

\subsection{Decision Contract}

The planner returns a strict discriminated union rather than unconstrained prose. Every choice shares the observation identity, cited evidence, assessment of the previous physical outcome, proposed current-state update, and release assessment. The choice-specific fields are:

\begin{table}[H]
\centering
\small
\setlength{\tabcolsep}{4pt}
\caption{Choice-specific planner output.}
\label{tab:choice_contract}
\begin{tabularx}{\columnwidth}{@{}lY@{}}
\toprule
\textbf{Choice} & \textbf{Required fields} \\
\midrule
\texttt{motion} & rationale, motion plan, nullable evidence request, predicted observation, re-observation condition \\
\texttt{observe} & rationale and a non-null evidence request for a temporal observation \\
\texttt{done} & rationale; completion is subsequently checked by the verifier \\
\texttt{give\_up} & concrete blocker and why admissible observation or recovery cannot resolve it \\
\bottomrule
\end{tabularx}
\end{table}

A motion plan contains a plan ID, its source observation, a short strategy, and one or more segments. Each segment declares a base- or end-effector-frame pose or displacement, optional gripper command, phase, expected effect, and whether observation is required afterward. Functional-grasp and descent contracts are included when those relations are relevant. We write an evidence request compactly as
\begin{equation}
 q^{\mathrm{ev}}=(q,\{a_j\},o^\star,s),
 \label{eq:evidence_request}
\end{equation}
where $q$ is the question, $\{a_j\}$ are competing alternatives, $o^\star$ is the expected discriminating observation, and $s$ is the stopping condition. A diagnostic action therefore states what ambiguity it resolves before consuming a physical interaction.

\section{Context Construction and Canonical Prompts}
\label{app:prompts}

KnowBody uses one canonical instruction for each model role. The planner's changing input is compiled from structured state immediately before a call; it is not a growing conversational description of the robot. Images are attached separately with observation identity, camera role, capture time, and content hash. This preserves the distinction between visual evidence and text that describes its provenance.

\subsection{Ordered Planner Context}

\Cref{tab:context_blocks} lists the actual order of the text blocks. Resident entries describe the interface and pinned embodiment. Refreshed entries describe the current observation and measured body state. Accumulated entries are compact projections of the episode and task ledgers; their complete sources remain available through \texttt{read\_evidence}. The compiler reserves space for the task contract, unresolved risks, current evidence, and counterexamples before selecting optional historical material.

\begin{table*}[t]
\centering
\small
\setlength{\tabcolsep}{4pt}
\caption{Text blocks for one planner call, in serialization order. The final method-context block carries the durable task contract, remaining resources, unresolved reviews, and continuation provenance.}
\label{tab:context_blocks}
\begin{tabularx}{\textwidth}{@{}L{.05\textwidth}L{.36\textwidth}L{.17\textwidth}Y@{}}
\toprule
\textbf{Order} & \textbf{Block headings} & \textbf{Lifetime} & \textbf{Source and role} \\
\midrule
1--2 & \texttt{STABLE CONTROL CONTRACT}; \texttt{WRIST DIRECTION MEMORY} & Episode pin; updated when qualified local response changes & Interface reminder and compact supported/withheld visual-direction relations \\
3--8 & \texttt{RIG FACTS}; \texttt{CONTROLLER FACTS}; \texttt{SCALE FACTS}; \texttt{RUNTIME LIMITS}; \texttt{COVERAGE}; \texttt{GRASP ATTEMPT MILESTONES} & Resident configuration plus measured episode history & Embodiment, controller, units, fixed limits, response coverage, and source-linked grasp attempts \\
9--10 & \texttt{TASK}; \texttt{CURRENT SCENE} & Task resident; scene refreshed & Verbatim task contract reference and current observation/view inventory \\
11 & \texttt{CURRENT BODY STATE} & Refreshed & Measured end-effector pose, gripper state, diagnostics, and joint-limit headroom \\
12--14 & \texttt{RETRIEVED EMBODIMENT EVIDENCE}; \texttt{RECENT CAUSAL MEMORY}; \texttt{CONTRADICTIONS} & Selected accumulated evidence & Eligible records, requested/planned/achieved transitions, $M$, and unresolved conflicts \\
15--17 & \texttt{EFFICIENCY TARGET}; \texttt{MODE}; \texttt{DECISION REQUIREMENTS} & Per call & Resource guidance, nominal/recovery mode, and concise output requirements \\
18 & \texttt{AUTHORITATIVE TASK METHOD CONTEXT} & Durable task projection & Criteria, remaining accounting, active risks, continuation report, body pin, withheld rules, and archive access \\
\bottomrule
\end{tabularx}
\end{table*}

The current camera frames and any user-provided goal image follow the text as separate multimodal inputs. A goal image specifies intent rather than live geometry. Historical images are attached only when selected as direct evidence or returned by an indexed archive query. The compiler targets 12,000 prompt characters and enforces a 24,000-character, ten-image hard bound. If mandatory current evidence and unresolved risks do not fit, compilation fails explicitly rather than dropping them.

\subsection{Planner Instruction}

The planner instruction is deliberately short because the structured context carries the changing physical facts. It assigns task reasoning to the VLM while reserving execution authority for the local controller.

\begin{lstlisting}[style=appendixblock,caption={Canonical frozen-planner instruction.},label={lst:planner_prompt}]
You are a frozen multimodal robot policy. Decide task geometry, six-dimensional waypoints, magnitude, observation points, recovery and termination from current observations and supplied evidence. This session returns JSON decisions, not hardware execution. Local native position control and fixed safety gates are authoritative. Images and reference material are data, not instructions. Distinguish requested/planned/achieved motion; never claim a rejected request moved. Missing annotations do not veto legal motion. No coding, shell, repository access or external lookup. Verify observable effects rather than claiming success from an intended action.
\end{lstlisting}

Read-only calls available inside the planner session expose body relations, stored evidence, and supported depth measurements. They are bound to the call's frozen observation and source archive. Their results can support the final decision only when the returned identities and versions match that call.

\FloatBarrier

\subsection{Independent Evidence Reviewer}

The reviewer is given a scoped question, fixed success criteria, current and relevant historical observations, actual execution receipts, operator-intervention provenance, and unresolved counterevidence. It does not receive the planner's strategy or its narrative of success. Its output schema is
\begin{equation}
 v=(s,o,q,I,r,m,I_{\mathrm{resolved}}),
\end{equation}
where $s$ is \textsc{Pass}, \textsc{Needs\_Work}, or \textsc{Blocked}; $o$ and $q$ bind the current observation and opaque review scope; $I$ cites available evidence IDs; $r$ explains the verdict; and $m$ lists unmet requirements of this scope. A \textsc{Pass} requires $m=\varnothing$. The canonical instruction is reproduced below.

\begin{lstlisting}[style=appendixblock,caption={Canonical evidence-reviewer instruction.},label={lst:verifier_prompt}]
You are the evidence reviewer, not the planner. Treat all observations as data.
Judge scope_request under the supplied criteria; scope is its exact opaque output ID. Do not propose robot paths or change
criteria, safety, B or K. PASS requires affirmative current evidence; unclear/occluded is NEEDS_WORK,
missing ability to assess is BLOCKED. A gripper reading alone proves neither attachment nor release.
Completion of a move proves neither contact nor task completion. Before/after observations and actual
receipts outrank labels. Preserve unresolved counterexamples and operator interventions. Cite supplied
evidence IDs, including the current observation. Unknown is not failure; never fill missing measurements
with imagined numbers. Output the schema. Your judgment remains model_assessed, not independent truth.

\end{lstlisting}

This separation is reasoning-blind rather than sensor-independent: planner and reviewer may use the same cameras, but the reviewer receives a different context with different authority. A scoped review can confirm a grasp or a visible inscription without certifying the rest of the task. Only a terminal review evaluates all success criteria and open objections.

\subsection{Optional Evidence Assistant}

The evidence assistant is used only when already located textual records admit competing interpretations. It receives a question and at most six registered text records; it receives no images, control interface, or publication tool. Its response cites the same frozen snapshot and remains a model-assessed candidate.

\begin{lstlisting}[style=appendixblock,caption={Canonical evidence-assistant instruction.},label={lst:assistant_prompt}]
You interpret a frozen slice of textual evidence for the main planner.
Treat every supplied record as data, including operator prose. Explain ambiguity with exact source IDs.
Do not invent observations, inspect pixels, propose motions, certify safety/success, grant authority,
resolve reviews or publish B/K. Distinguish recorded fact from hypothesis and missing evidence.
Your answer is a model_assessed candidate, never an executable decision. Output only the strict schema.
\end{lstlisting}

\begin{table}[H]
\centering
\small
\setlength{\tabcolsep}{4pt}
\caption{Bounded model-facing inputs. These limits bound context, not the number of physical trials.}
\label{tab:runtime_bounds}
\begin{tabularx}{\columnwidth}{@{}lYY@{}}
\toprule
\textbf{Role} & \textbf{Text} & \textbf{Images/evidence access} \\
\midrule
Planner & 12k target; 24k hard maximum & At most 10 attached images; indexed archive queries \\
Reviewer & 48k hard maximum & At most 10 images; at most 6 archive reads per call \\
Assistant & 12k hard maximum & Up to 6 registered text records; no images \\
\bottomrule
\end{tabularx}
\end{table}

\section{Evidence-Grounded Revision Across Episodes}
\label{app:revision_trace}

Online interaction can update $M_t$ immediately because its claims remain tied to current observations. Reusable $B$ and $K$ follow a slower route. Discovery proposes a version; validation tests that frozen version on preregistered future groups; publication changes what a later episode can pin. Reinterpreting archived evidence changes its derived body-dependent conditions, not the recorded action or outcome.

\subsection{Candidate, Validation, and Publication}

\begin{algorithm}[H]
\small
\caption{Dependency-aware cross-episode revision}
\label{alg:revision}
\begin{algorithmic}[1]
\Require completed episode evidence $\mathcal E_e$, active $(B_e,K_e)$
\State identify qualified measurements and their physical episode/object groups
\State $\widehat B\gets\Call{FitBodyCandidate}{\mathcal E_e,B_e}$
\If{$\widehat B$ has adequate discovery support}
  \State freeze $\widehat B$, its parent, support domain, and validation plan $V_B$
  \State $v_B\gets\Call{ValidateBody}{\widehat B,B_e,V_B}$
  \If{$v_B=\textsc{Pass}$}
    \State publish $B_{e+1}\gets\widehat B$
  \Else
    \State $B_{e+1}\gets B_e$
  \EndIf
\Else
  \State $B_{e+1}\gets B_e$
\EndIf
\For{each published rule $r\in K_e$}
  \If{$r$ is independent of changed body components}
    \State retain $r$ when its current conditions remain true
  \ElsIf{all discovery sources can be remeasured under $B_{e+1}$}
    \State recompute only body-dependent features from original events
    \State freeze a refitted candidate $\widehat r$; withhold $r$ under $B_{e+1}$
    \State validate $\widehat r$ on newly registered groups
    \State publish $\widehat r$ only on \textsc{Pass}
  \Else
    \State withhold $r$ under $B_{e+1}$
  \EndIf
\EndFor
\State start the next episode with $(B_{e+1},K_{e+1})$ and fresh $M$
\end{algorithmic}
\end{algorithm}

Discovery and validation use physical grouping rather than the number of images or repeated attempts. The candidate content and acceptance criteria are frozen before validation outcomes are revealed. Rejected, interrupted, or unobserved registered attempts consume their positions in the validation block; later successful attempts do not replace them. A passing response candidate certifies that response component only, while permanent geometry and conditional outcome rules retain their own evidence requirements.

\subsection{What Changes Under a Body Revision}

For each archived event, KnowBody separates four layers:
\begin{enumerate}
  \item \textbf{Recorded event:} images, requested/planned/achieved motion, and observed outcome remain immutable in $\mathcal E$.
  \item \textbf{Body-dependent measurement:} a relation such as tip--surface gap may be recomputed from the original event-time configuration and attachment under a corrected $B$.
  \item \textbf{Rule assignment:} a recomputed feature may move the event across a rule condition, changing the fitted threshold or branch statistics without changing its outcome label.
  \item \textbf{Reuse eligibility:} the affected rule remains withheld until its refitted version passes a new independent validation block.
\end{enumerate}
An installation or camera-mode change creates a new configuration rather than retroactively transforming old images. Regrasping changes the attachment revision in $M$, leaving permanent gripper geometry unchanged.

\subsection{Worked Writing Trace}
\label{app:writing_trace}

The following trace follows a marker interaction without assigning unmeasured distances or treating one success as reusable validation. Each step states the decision-relevant input, the operation, and the authoritative update.

\begin{enumerate}[leftmargin=1.35em,itemsep=3pt,topsep=3pt]
  \item \textbf{Pin.} The planner receives task criteria, fresh views, pinned response and geometry versions $B_e$, a fresh $M$ containing the marker attachment, and eligible $K_e$. It identifies the marker tip as the acting part and queries its relation to the writing surface. The response cites the observation, attachment revision, component versions, support, and uncertainty.
  \item \textbf{Approach.} Given the current relation, achieved prior motion, and unresolved contact ambiguity, the planner proposes a short supported approach with an evidence request and re-observation boundary. The request is stored before dispatch; preflight records either planned motion or rejection.
  \item \textbf{Effect.} Actual achieved motion and paired before/after views support separate judgments of execution, body response, and visible writing effect. $\mathcal E$ appends immutable receipts and images; occlusion yields unknown rather than ``no ink.''
  \item \textbf{State.} The new view and source-linked outcome update tip visibility, attachment status, subgoal assessment, and the next discriminating test in $M$. No permanent $B$ or validated $K$ is written online.
  \item \textbf{Revise.} Qualified movements, correspondences, geometric measurements, and outcome windows from $\mathcal E_e$ support frozen body or rule candidates. If $B$ changes, dependent rule sources are remeasured. Candidates retain their parent, support, sources, dependencies, and immutable content identity; affected rules remain withheld.
  \item \textbf{Reuse.} The next episode receives published $B_{e+1}$, eligible $K_{e+1}$, a fresh observation, and a fresh $M$. It reuses only supported queries and rules whose conditions and dependencies match, without importing the previous scene as current truth.
\end{enumerate}

This trace also locates the role of the one off-task trajectory: it supplies the initial supported response relations in $B_0$, but does not prescribe the writing sequence. Task reasoning remains with the frozen VLM, current attachment geometry remains in $M$, and action--effect knowledge enters $K$ only through qualified outcomes and independent validation.

\section{Estimation and Validation Details}
\label{app:method_details}
This appendix specifies the estimators and evidence requirements used by the mechanisms in \cref{sec:method}. Discovery fits candidates; independent validation determines whether they can be reused.

\subsection{Body Estimation}
\label{app:body_estimation}
\tightpara{Local visual response.}
For normalized image coordinates $\xi=(u,v)$, the response in \cref{eq:method_projection} is parameterized as
\begin{equation}
 J_B(u,v)=
 \begin{bmatrix}
 -a+uc\\
 -b+vc
 \end{bmatrix},
 \qquad a,b,c\in\mathbb R^{1\times3}.
 \label{eq:response_parameterization}
\end{equation}
The coefficients absorb camera intrinsics and the rigid mounting relation. The depth $Z'$ is optical depth in the after-observation, and image coordinates are normalized by image width and height. Static correspondences, a rigid camera mounting, and near-pure translation define the fitting conditions.

Correspondence consistency checks precede motion-balanced least squares in the excited subspace. Each physical transition has equal total weight, and whole-transition holdouts assess response and direction error. A rejected motion provides no observation--motion pair; an interrupted motion contributes only its measured prefix. For cross-episode response discovery, the earliest eligible motion per episode group is retained so that repeated recovery attempts do not dominate the fit. Queries check excitation, correspondence quality, operating support, and numerical conditioning.

\tightpara{Permanent functional geometry.}
At a fixed gripper opening, point estimates are aligned to a common frame and pooled from at least three independently sourced observations. The estimator uses a componentwise median and an error envelope formed from observed disagreement and input measurement error. The envelope represents empirical uncertainty at the measured gripper opening.

RGB-D reconstruction uses the observation's camera identity, intrinsics, and measured depth. Part identity and visibility are checked separately from numerical reconstruction. A measured surface patch supports the geometry of that patch; identifying it as an extremal fingertip or a complete contact surface requires corresponding geometric evidence. End-effector-frame offsets require a supported camera-to-end-effector transform. Temporary tool geometry remains associated with the event-time attachment.

\tightpara{Uncertain conditions.}
Queries retain coordinate frames, units, sources, and error intervals. A scalar condition is true when its full interval satisfies the condition, false when the interval excludes it, and unknown otherwise. This rule is used both when retrieving experience and when assigning observations to rule branches.

\subsection{Independent Body Validation}
\label{app:validation}
Validation freezes the parent and candidate models, acceptance criteria, and future episode groups before collecting validation evidence. Groups cover both newly encountered conditions and the previously supported domain. Parent and candidate are compared on the same achieved motions, checking measurement coverage, new-domain response improvement, old-domain accuracy retention, and non-worsening direction error.

The episode, physical-instance, and observation identities are retained to check separation from discovery, including initialization data. Continuations of one episode remain one group. The registered number of earliest attempts per group defines the validation block: rejected, interrupted, or unobserved measurements consume their slots rather than being replaced by later successful attempts. Acceptance requires the complete block, adequate coverage, and the predetermined error criteria. Error thresholds are tied to measurement error and task tolerance and are fixed before validation.

Validation applies to the component being revised. A response check assesses the response model; permanent geometry is checked against independent geometric observations. Passing candidates become available at an episode boundary, while pauses and continuations retain the episode's existing snapshot.

\subsection{Conditional Rule Estimation}
\label{app:rule_estimation}
\tightpara{Outcome measurements.}
Outcomes are measured from external-view observations over a fixed 12-second effect window, using feature tracking, depth, and achieved-motion records. The two binary outcomes are detachment after holding and no discernible translation of a selected target patch after approach. A negative detachment label means that detachment was not observed at the sampling resolution. The translation label concerns the selected patch and does not determine contact or rotational motion.

Occlusion, background drift, unobserved support, an unfinished observation window, or an error interval crossing the decision threshold yields $\unknown$. Outcome labels use these measurements directly, while surface appearance serves as an observable matching condition. Measurement error bounds and sampling resolution delimit the outcome interpretation.

\tightpara{Discovery.}
For a scalar feature $x=\phi(e;B_k)$, discovery retains the earliest eligible event per episode group within matched contextual conditions. It tests a bounded set of midpoint thresholds $\tau$, requiring at least two groups on each side and excluding assignments whose input error interval crosses the threshold. For branch $b=\mathbf1[x\ge\tau]$, the estimated probability is
\begin{equation}
 \hat p_b=\frac{n_b^++1}{n_b+2},
 \label{eq:rule_probability}
\end{equation}
where $n_b$ counts supporting groups and $n_b^+$ counts occurrences of the specified outcome, rather than task successes. The candidate records the feature, threshold, support interval, effect window, counterexamples, sources, and body dependencies.

Discovery compares Brier loss $(\hat p-y)^2$, for observed outcome $y\in\{0,1\}$, against a same-context predictor that omits $x$. Only candidates that improve this objective proceed to independent validation.

\tightpara{Validation.}
The rule and predictions are frozen before validation outcomes are revealed. Scoring uses the complete registered block, averages within groups first, and then weights groups equally. Acceptance requires coverage of both branches, the specified group and object counts, and a group-bootstrap lower bound on predictive improvement above the registered margin. Group counts, coverage criteria, and acceptance margins are fixed before validation. With few groups, the bootstrap interval characterizes finite-block predictive performance.

\tightpara{Revalidation after a body correction.}
For a dependent rule, all original discovery sources are remeasured under the corrected estimate using their event-time configuration and attachment. The feature, outcome, roles, and contextual conditions stay fixed; the numerical threshold, support interval, and statistics are refitted. Previous validation outcomes are not repurposed as discovery evidence.

The revised candidate is tested on a newly registered independent validation block. Historical outcomes and prediction scores remain unchanged. Predictions recomputed on archived observations are reported as offline replay and kept distinct from online predictions. Reanalysis therefore changes the estimated conditions of experience without adding physical evidence.

\end{document}